\documentclass[sigconf]{aamas} 

\usepackage{balance} 

\usepackage{amsmath,amsfonts} 
\usepackage{algorithmic}
\usepackage{graphicx}
\usepackage{textcomp}
\usepackage{xcolor}
\usepackage{bm}
\usepackage{bbm}
\usepackage{todonotes} 
\usepackage{multirow}
\usepackage{booktabs}
\usepackage{algorithm}
\usepackage{microtype}
\usepackage{eqparbox}
\usepackage{url}
\usepackage{subcaption}
\usepackage{relsize}
\usepackage{hyperref}

\usepackage{xcolor} 
\usepackage{tikz}
\usetikzlibrary{fit,calc}

\colorlet{pink}{red!40}
\colorlet{blue}{cyan!60}

\renewcommand{\algorithmiccomment}[1]{\bgroup\hfill$\triangleright$~#1\egroup}

\newcommand{\argmax}[1]{\underset{#1}{\operatorname{arg}\,\operatorname{max}}\;}

\setcopyright{ifaamas}
\acmConference[AAMAS '21]{Proc.\@ of the 20th International Conference on Autonomous Agents and Multiagent Systems (AAMAS 2021)}{May 3--7, 2021}{Online}{U.~Endriss, A.~Now\'{e}, F.~Dignum, A.~Lomuscio (eds.)}
\copyrightyear{2021}
\acmYear{2021}
\acmDOI{}
\acmPrice{}
\acmISBN{}

\acmSubmissionID{652}

\title[Action Advising with Advice Imitation in Deep Reinforcement Learning]
{Action Advising with Advice Imitation\\in Deep Reinforcement Learning}

\author{Erc\"{u}ment \.{I}lhan}
\affiliation{
  \department{School of Electronic Engineering and Computer Science}
  \institution{Queen Mary University of London}
  \city{London, UK}}
\email{e.ilhan@qmul.ac.uk}

\author{Jeremy Gow}
\affiliation{
  \department{School of Electronic Engineering and Computer Science}
  \institution{Queen Mary University of London}
  \city{London, UK}}
\email{jeremy.gow@qmul.ac.uk}

\author{Diego Perez-Liebana}
\affiliation{
  \department{School of Electronic Engineering and Computer Science}
  \institution{Queen Mary University of London}
  \city{London, UK}}
\email{diego.perez@qmul.ac.uk}

\begin{abstract}
Action advising is a peer-to-peer knowledge exchange technique built on the teacher-student paradigm to alleviate the sample inefficiency problem in deep reinforcement learning.
Recently proposed student-initiated approaches have obtained promising results.
However, due to being in the early stages of development, these also have some substantial shortcomings.
One of the abilities that are absent in the current methods is further utilising advice by reusing, which is especially crucial in the practical settings considering the budget constraints in peer-to-peer interactions.
In this study, we present an approach to enable the student agent to imitate previously acquired advice to reuse them directly in its exploration policy, without any interventions in the learning mechanism itself.
In particular, we employ a behavioural cloning module to imitate the teacher policy and use dropout regularisation to have a notion of epistemic uncertainty to keep track of which state-advice pairs are actually collected.
As the results of experiments we conducted in three Atari games show, advice reusing via imitation is indeed a feasible option in deep RL and our approach can successfully achieve this while significantly improving the learning performance, even when it is paired with a simple early advising heuristic.
\end{abstract}

\keywords{Deep Reinforcement Learning; Deep Q-Networks; Action Advising}

\newcommand{\BibTeX}{\rm B\kern-.05em{\sc i\kern-.025em b}\kern-.08em\TeX}

\begin{document}


\pagestyle{fancy}
\fancyhead{}


\maketitle 

\section{Introduction}

Deep reinforcement learning (RL) has made it possible to build end-to-end learning agents without having to handcraft task-specific features, as it is showcased in various challenging domains such as StarCraft II \cite{DBLP:journals/corr/abs-1708-04782} and DotA II \cite{DBLP:journals/corr/abs-1912-06680} in the recent years.
These feats make deep RL a great candidate to be employed in complex real-world sequential decision-making problems.
However, achieving the reported levels of performance usually requires millions of environment interactions due to the deep learning induced complexity as well as the exploration challenges in RL itself.
Even though this may seem negligible in most of the experimental domains considering the immense amount of computing power available to be utilised through parallel simulations, it usually poses a problem in the real-world scenarios due to the interaction costs and safety concerns.
Furthermore, since RL is an inherently online learning approach, it is desired for the agents to be continually learning after they have been deployed too.
For these reasons, it is crucial to improve sample efficiency in deep RL, which is actively investigated in several lines of research.
One promising approach to tackle this setback is leveraging some legacy knowledge acquired from other entities such as agents, programs or humans.

Peer-to-peer knowledge transfer in deep RL has been investigated in various forms to this date \cite{DBLP:journals/aamas/SilvaWCS20}.
A popular approach, namely Learning from Demonstrations (LfD), focuses on incorporating a previously recorded dataset in the learning process.
By taking some dataset generated by another competent \cite{DBLP:conf/aaai/HesterVPLSPHQSO18} or imperfect \cite{DBLP:conf/iclr/GaoXLYLD18} peer, the learning agent tries to make the most out of the available information through off-policy learning and extra loss terms.
Another promising, yet under-investigated class of techniques, namely Action Advising \cite{DBLP:conf/atal/TorreyT13}, aims to take advantage of a competent peer interactively when there is no pre-recorded data.
The learning agent acquires advice in the form of actions from a teacher for a limited number of times defined by a budget that resembles the practical limitations of communication and attention.
This approach is especially beneficial in the situations where there is no way to access the actual task before the online training, data collection is costly or the relevant data that will do the most contribution in the learning can not be determined.
Action advising methods in deep RL today are quite limited and therefore have several shortcomings.
An important one of these as we address in this study is not being able to make further use of the advice beyond its collection.

The scope of the action advising problem is generally limited to answering ``\emph{when} to ask for advice?''.
It is commonly not of any interest how the collected advice is utilised by the student agent's task-level RL algorithm, e.g., how it is stored, replayed, discarded; especially since these are dealt with by the studies that focus on off-policy experience replay dynamics in general \citep{DBLP:journals/corr/SchaulQAS15, de2015importance}, or the specific case of having demonstration data as in LfD.
However, even without interfering with the student's task-level learning mechanism, it is still possible to make more of advice through reuse.
Current action advising algorithms in deep RL have no way of telling if they have asked for advice in a very similar or even identical state already in the learning session.
Thus, they do not record these in any way, and usually end up requesting advice from the teacher redundantly.
In order to address this, we incorporate a separate neural network to do behavioural cloning \citep{DBLP:journals/neco/Pomerleau91} on the samples (state-action pairs which are equal to the state-advice pairs in the context of action advising) collected from the teacher.
This network then will be able to serve as a state-conditional generative model that will let us sample advice for any given observation.
However, since this model should also have a notion of distinguishing the recorded states from the unrecorded ones to avoid producing false advice for unfamiliar states, we also propose incorporating a well known regularisation mechanism called Dropout \citep{DBLP:journals/jmlr/SrivastavaHKSS14} within this network to serve as an epistemic uncertainty estimator \citep{DBLP:conf/icml/GalG16} which will allow the student to determine whether the state is recorded by comparing this estimation with a threshold.

Our contributions in this study are as follows:
First, we show that it is possible to generalise teacher advice across similar states in deep RL with high accuracy.
Second, we present a RL algorithm-agnostic approach to memorise and imitate the collected advice that is suitable with the deep RL settings.
Finally, we demonstrate that advice reuse via imitation provides significant boosts in the learning performance in deep RL even when it is paired with a simple baseline like early advising.

\section{Related Work} \label{sec:related_work}

The majority of action advising studies to date have been conducted in classical RL settings.
\cite{DBLP:conf/atal/TorreyT13} was the first study to formalise action advising within a budget constrained teacher-student framework.
Specifically, they studied the teacher-initiated scenario and came up with several heuristics to distribute the advising budget to maximise the student's learning performance, such as early advising and importance advising.
This work was then extended to introduce several new state importance metrics \cite{DBLP:journals/connection/TaylorCFVT14}.
In \cite{zimmer2014teacher}, action advising problem was approached as a meta-RL problem itself.
Instead of relying on heuristics, the authors attempted to learn the optimal way to distribute advising budget by using a measurement of the student's learning acceleration as the meta-level reward.
Besides these studies that only consider the teacher's perspective, \cite{DBLP:conf/ijcai/AmirKKG16} explored student-initiated and jointly-initiated variants considering the impracticality of requiring the teacher's attention constantly.
They achieved results on-par with the previous work without requiring the teacher full-time.
\cite{DBLP:conf/ijcai/ZhanBT16} shed light in the theoretical aspects of action advising problem by using a more general setting involving multiple teachers and demonstrated the effects of having good or bad teachers.
In \cite{DBLP:conf/atal/SilvaGC17}, the authors adopted the teacher-student framework in cooperative multi-agent RL where the agents learn from scratch and hold no assumptions of their teacher roles and expertise.
By proposing state counting as a new heuristic in this setting, they successfully accelerate team-wide learning of independent learners.
More recently, learning to teach concepts was further investigated in \cite{DBLP:journals/make/FachantidisTV19} with a focus on the properties that make for a good teacher.
In this work, besides learning \emph{when} to advise, the teachers also learn \emph{what} to advise.
Similarly, \cite{DBLP:conf/aaai/OmidshafieiKLTR19} adopted the meta RL approach, this time as a deep RL scale.
They considered a team of two agents that learn to cooperate from scratch in tabular multi-agent tasks.
\cite{DBLP:conf/atal/ZhuCLH20} is one of the most recent studies conducted in tabular settings.
The idea of reusing the previously collected advice in order to make the most out of a given small budget was studied.
By devising several heuristics to serve as reusing schedules, they demonstrated promising results that outperform the algorithms incapable of advice reusing.

The domain of deep RL is a fairly new area for action advising where the primary choice is the student-initiated approaches.
\cite{DBLP:journals/corr/abs-1812-02632} is one of the first studies to explore the idea of action advising in deep RL.
They combined the LfD paradigm \cite{DBLP:conf/aaai/HesterVPLSPHQSO18} with interactive advice exchange under the name of \emph{active learning from demonstrations} to collect demonstration data on-the-fly to be utilised via imitation capable loss terms as used in \cite{DBLP:conf/aaai/HesterVPLSPHQSO18}.
Furthermore, they proposed using epistemic uncertainty estimations of the student agent's model to time this advice. 
Later, \cite{DBLP:conf/atal/Kim0OLRHTMCH20} was proposed as an extension of \cite{DBLP:conf/aaai/OmidshafieiKLTR19}.
This time, meta deep RL to address learning to teach idea was applied in the problems that are deep RL in the task-level.
Through multiple centralised learning sessions, agents in a set of cooperative multi-agent tasks were made to learn taking student and teacher roles as needed in order to improve team-wide knowledge.
To do so, they adopted \emph{hierarchical reinforcement learning} \citep{DBLP:conf/nips/NachumGLL18} to deal with the meta-level credit assignment problem of the teacher actions.
In \cite{DBLP:conf/cig/IlhanGP19}, the formal action advising framework was scaled up to deep RL level for the first time.
Similarly to \cite{DBLP:conf/atal/SilvaGC17}, a team of agents in a cooperative multi-agent scenario were made to exchange advice by embracing teacher or student roles as needed.
This was accomplished by using \emph{random network distillation} (RND) \cite{DBLP:journals/corr/abs-1810-12894} to replace state counting with state novelty, hence introducing a new heuristic that is applicable in non-linear function approximation domain.
Later on, \cite{DBLP:conf/aaai/SilvaHKT20a} proposed the idea of uncertainty-based action advising as in \cite{DBLP:journals/corr/abs-1812-02632}, though without employing any additional loss terms.
To access uncertainty estimations, they studied the case of student agent with a multi-headed network architecture in particular.
In a more recent work \cite{ilhan2020studentinitiated}, student-initiated scenario is further studied to devise a more robust heuristic able to handle extended periods of absence of teacher as well as having no requirements in the student's task-level architecture by completely decoupling the module that is responsible for advice timing from the student's model.
Even though this method also uses the state novelty heuristic proposed in \cite{DBLP:conf/cig/IlhanGP19}, they operated on the advised states directly rather than every encountered state.

Clearly, none of the related work in deep RL addressed further utilisation of collected advice, besides \cite{DBLP:journals/corr/abs-1812-02632} which does it through interfering with the student's learning mechanism (via a custom loss function), unlike our approach.
The study that is closest to the idea we present in this paper is \cite{DBLP:conf/atal/ZhuCLH20}; though, it is limited to the tabular RL domains only.
Such a setting makes it more straightforward for the agent to precisely memorise the state-advice pairs in a look-up table to be able to reuse anytime.
Furthermore, the executed advice usually has an instantaneous impact on the agent behaviour in the case of tabular RL, which presents unique options to assess their usefulness.
Since these advantages are absent in deep RL, our work deals with different challenges than those in \cite{DBLP:conf/atal/ZhuCLH20}.
\section{Background} \label{sec:background}

\subsection{Reinforcement Learning} \label{subsec:rl}

Reinforcement Learning (RL) \cite{sutton2018reinforcement} is a trial-and-error learning paradigm that deals with sequential decision-making problems where the environment dynamics are unknown.
In RL, Markov Decision Process (MDP) formalisation is used to model the environment and the interactions within.
According to this, an environment is defined by a tuple $\langle \mathcal{S}, \mathcal{A}, \mathcal{R}, \mathcal{T}, \gamma \rangle$ where $\mathcal{S}$ is the finite set of states, $\mathcal{A}$ is the finite set of actions, $\mathcal{R} \colon \mathcal{S} \times \mathcal{A} \times \mathcal{S} \rightarrow \mathbb{R} $ is the reward function, $\mathcal{T} \colon \mathcal{S} \times \mathcal{A} \rightarrow \Delta(\mathcal{S}) $ defines the state transitions and $\gamma \in [0,1]$ is the discount factor.
The agent to interact within an environment receives a state observation $s_t$ at each timestep $t$, and executes an action $a_t$ to advance to the next state $s_{t+1}$ while obtaining a reward $r_t$.
Actions of the agent are determined by its policy $\pi\colon S \rightarrow A$, and the agent's objective is to construct a policy that maximises the expected sum of discounted rewards in any timestep, which can be formulated as $\sum_{k=0}^{T} \gamma^{k} r_{t + k}$ for a horizon of $T$ timesteps.


\subsection{Deep Q-Networks} \label{subsec:dqn}

Deep Q-Network (DQN) \cite{DBLP:journals/corr/MnihKSGAWR13} is a prominent RL algorithm that tries to obtain the optimal policy in complex domains by employing non-linear function approximation via neural networks to learn mapping any given state into state-action values ($Q(s,a)$).
Specifically, a neural network $G_{\theta}$ with randomly initialised weights $\theta$ is trained over the course of learning to minimise the loss $(r_{k+1} + \gamma \max_{a'} Q_{\bar{\theta}}(s_{k+1}, a') -  Q_{\theta}(s_{k}, a))^2$ with batches of transitions that are collected on-the-fly and stored in a component called replay memory.
Periodically using the samples from this memory, which is referred to as \emph{experience replay}, is an essential mechanism in DQNs.
As well as improving sample efficiency by reusing samples multiple times, it also breaks the non-i.i.d. property of sequentially collected data.
Furthermore, DQNs also employ another trick to aid convergence.
Since both the Q-value targets and network weights are learned at the same time, there is a significant amount of non-stationarity seen in these target values used in the loss function, which may introduce further instabilities due to the bootstrapped updates.
In order to alleviate this, a separate copy of $G$ is held with weights $\bar{\theta}$ that are updated periodically with copies of $\theta$, to be used in the target term in the loss function.

Due to its end-to-end learning and discarding the need for hand-crafted features, DQN has become a very popular approach in the field of RL that is followed by further enhancements over the years.
The most substantial ones among these are identified and combined in a version called Rainbow DQN \cite{DBLP:conf/aaai/HesselMHSODHPAS18}.
In our study, we employ double Q-learning \cite{DBLP:conf/aaai/HasseltGS16} and dueling networks \cite{DBLP:conf/icml/WangSHHLF16} among these essential modifications.

\subsection{Behavioural Cloning} \label{subsec:behavioural_cloning}

Behavioural cloning \citep{DBLP:journals/neco/Pomerleau91} refers to the ability of imitating a demonstrated behaviour.
It is especially useful in the situations where it is more difficult to specify reward functions than to provide some expert demonstration.
The simplest way of achieving this in the domain of deep RL is to train a non-linear function approximator, e.g. neural network $G_{\omega}$ with weights $\omega$, through supervised learning on the provided demonstration samples in the form of state-action pairs denoted by $\langle s, a \rangle$.
This is done by treating these as i.i.d. samples and minimising an appropriate loss function such as $\mathcal{L} (\omega) = \sum_{(s,a) \in D}^{} -log G_{\omega}(a \mid s)$.
Consequently, a state-conditional generative model is obtained that is capable of imitating the expert actions for the demonstrated states.
In practice, however, this approach is unreliable to be used as a task policy as it is.
This is because the agent often encounters states that are not contained in the provided dataset, and therefore, end up exhibiting sub-optimal behaviour in these states which lead to further divergence in the trajectories.
However, adopting the idea in this most basic form is sufficient in our study as it provides us the adequate functionality of generating actions correctly for the states we ensure $G_{\omega}$ is trained with.

\subsection{Dropout} \label{subsec:dropout}

Dropout \cite{DBLP:journals/jmlr/SrivastavaHKSS14} is a simple yet powerful regularisation method developed to prevent neural networks from overfitting.
Its working principle is based on involving some random noise in the hidden layers of the networks.
A neural network layer with the feed-forward operation can be described as $\boldsymbol{y} = f(\boldsymbol{w} \boldsymbol{x} + b)$, where the output is $\boldsymbol{y} \in \mathbb{R}^q$, the input is $\boldsymbol{x} \in \mathbb{R}^{p}$, the network weights for this particular layer are $\boldsymbol{w} \in \mathbb{R}^{q \times p}$ and $b \in \mathbb{R}^{q}$, $f$ is any activation function, for input size of $p$ and output size of $q$.
In a layer with dropout, this equation takes the form of $\boldsymbol{y} = f(\boldsymbol{w} \tilde{\boldsymbol{x}} + b)$ where $\tilde{\boldsymbol{x}} = \boldsymbol{r} * \boldsymbol{x}$ represent randomly dropped out input which is determined by $r \sim Bernoulli (p)$.
Hence, the learning process gets to be regularised with this random noise which is re-determined in every forward pass.
The value $p$ controls the rate of dropout and is responsible for the regularisation strength.

In addition to its regularisation capability, dropout can also be used to estimate epistemic uncertainty of a neural network model, as shown in \cite{DBLP:conf/icml/GalG16}.
For any particular input, performing forward passes multiple times yield different outputs due to the dropout induced stochasticity, which can be treated as an approximation of probabilistic deep Gaussian process.
Following this idea, the variance in these output values can therefore be interpreted as a representation of the model's uncertainty.
Finally, since these forward passes can be performed concurrently, this approach provides a practically viable option to evaluate the uncertainty in deep learning models.

\subsection{Action Advising} \label{subsec:action_advising}

Action advising \cite{DBLP:conf/atal/TorreyT13} is a knowledge exchange approach built on the teacher-student paradigm.
Requiring only a common set of actions and a communication protocol between the teacher and the student makes this a very flexible framework.
In its originally proposed form, the learning agent (student) is observed by an experienced peer (teacher) and is given action advice to be treated as high quality explorative actions to accelerate its learning.
However, maximum number of these interactions are limited with a budget constraint considering the real-world conditions where communication and attention span are usually limited.
Therefore, the approaches that adopt this idea address the question of \emph{when} to exchange advice in order to maximise the learning performance.
This is usually accomplished either by performing meta-learning over multiple learning sessions or by following heuristics as we do in this study.

Currently, there are several heuristic approaches with varying complexities and advantages in the deep RL domain such as early advising, random advising, uncertainty-based advising and novelty-based advising.
In this paper, we incorporate early advising as the baseline to build our method on.
Despite its simplicity, this method performs very well in deep RL especially with small budget scenarios \cite{ilhan2020studentinitiated}.
This is because the earlier samples have far more impact on the learning in deep RL models since providing high quality transitions that contains rewards provide more stable Q-value targets early on which can significantly reduce the non-stationarity in the learning process.
Finally, since the teacher is followed consistently in this approach, the student is more likely to encounter the critical states that would require deep exploration. This is an important property to have when it comes to spending the budget wisely.
\section{Proposed Approach} \label{sec:proposed_approach}

We follow the standard MDP formalisation given in Section~\ref{subsec:rl} in our problem definition.
In this setting, a student agent that employs an off-policy deep RL algorithm performs learning in an episodic single-agent environment through trial-and-error interactions.
It receives an observation $s_t$ and then executes an action $a_t$ generated by its policy $\pi_S$ to receive a reward $r_t$ at each timestep $t$, in order to maximise its cumulative discounted rewards in any episode.
According to the teacher-student paradigm (Section~\ref{subsec:action_advising}) we adopt, there is also an isolated peer that is competent in this same task, and is referred to as the teacher.
For a limited number of times defined by the action advising budget $b$, the student is allowed to acquire an action advice from the teacher for the particular state $s$ it is in.
While the teacher can have its own teaching strategies to generate actions to advise, in our setting, we determine the action to be advised greedily from the teacher's behaviour policy as $\pi_T(s)$.
This is a commonly followed approach with the assumption of the teacher and the student's optimal task-level strategies are equivalent.
The student considers this advice as a part of a high-reward strategy and follows them upon collection.
In this final form of the problem, the student's objective is to spend its budget at the most appropriate times to maximise its learning performance.

We aim to devise a method that will enable the student to memorise the collected advice to be able to re-execute them in the similar states; therefore, avoiding wasting its budget in redundant states and potentially being able to follow the teacher advice many more times than its budget.
In tabular RL, this is trivial to achieve simply by storing the advised actions paired with the states in a look-up table.
When it comes to deep RL where any particular observation is not expected to be encountered more than once, however, there needs to be a generalisable approach.
For this purpose, we propose the student agent to employ a separate behavioural cloning module, which consists of a neural network as the state-conditional generative model $G_{\omega} \colon \mathcal{S} \rightarrow \mathcal{A}$.
By training $G_{\omega}$ in a supervised fashion with the obtained state-advice pairs (stored in a buffer $D$) to minimise the negative log-likelihood loss $\mathcal{L} (\omega) = \sum_{(s,a) \in D}^{} -log G_{\omega}(a \mid s)$, the student can imitate the teacher's advice to reuse them accordingly.
However, this method does not have any mechanisms to prevent the student from generating incorrect advice from the states it has not collected.
Therefore, we also employ Dropout regularisation in $G_\omega$ in order to grant this behavioural cloning module a notion of epistemic uncertainty through measuring the variance in the outputs obtained from multiple forward passes for a particular input state.
We denote this uncertainty estimation by $G_{\omega}^{\mu}(s)$.
The states $G_{\omega}$ is trained on will be less susceptible to the variance caused by the dropout and yield smaller uncertainty values.
By this means, the student can determine how likely a state is to be already recorded as advised when it comes to reusing them, and can make a decision according to a threshold.

An obvious question regarding the feasibility of reusing advice in deep RL arises here: can the teacher's advice be generalised over similar states accurately?
As we investigate in the experiments in Section~\ref{sec:results_and_discussion}, actions generated by the teacher policy usually span over similar states.
Clearly, the uncertainty threshold to consider a state as recorded is responsible for the trade-off between the reusing amount and the accuracy of the self-generated teacher advice.
A small threshold value makes the student reuse its budget in fewer states with higher accuracy, whereas a larger value results in more frequent reusing with lower accuracy.

The detailed breakdown of our approach is summarised with an emphasis on the proposed modifications as follows (as also shown in Algorithm~\ref{alg:proposed_approach}):
The student starts with a randomly initialised $G_{\omega}$ and empty $D$.
At each timestep $t$ with the (observed) state $s_t$ and an undecided action $a_t$, the student first checks if $D$ has any new samples.
As soon as $D$ reaches the size defined by $n_{D}$, $G_{\omega}$ is trained with mini-batch gradient descent over the samples in $D$ for $k_{bc}$ iterations.
Afterwards, if the environment was reset (a new episode started), the student determines whether to enable advice reuse via imitation for this particular episode with a probability of $\varepsilon_{reuse}$, which is combined with other conditions too later on in the algorithm.
The idea behind employing this condition is to ensure that the student can also execute its own exploration policy in order to increase the data diversity in its replay memory, which is crucial to improve the quality of learning.
Furthermore, determining this variable on an episodic basis lets the agent follow consistent policies in the exploration steps, rather than dithering between two policies.
In the next phase, the student deals with the advice collection.
We adopt the simple yet strong baseline of early advising here.
According to this, the agent just collects advice without any conditions until its budget runs out.
In the next phase, the student decides whether to reuse advice generated by its $G_{\omega}$.
There are several conditions to be satisfied for this to occur in addition to the advice reuse being allowed for this particular episode.
Firstly, $a_t$ must be non-determined, which implies the agent has not collected any advice from the teacher already.
Secondly, $G_{\omega}$ must be already trained, so that it can generate meaningful actions.
Then, the student also checks if its own action $\pi_{S}(a \mid s_{t})$ is explorative.
This condition limits the action advising actions to the exploration steps only in order to prevent overriding the student's actual policy which may result in lack of Q-value corrections and cause deteriorative effects when too much advising occurs.
Finally, it is checked whether $G_{\omega}^{\mu}(s_t)$ is smaller than the reuse threshold $\tau_{reuse}$.
Incorporating such threshold is important to limit the imitated advice to the states that have low uncertainty according to $G_{\omega}$ to achieve higher accuracy of generating correct teacher actions.
On one hand, having this threshold too high would make the student consistently follow $G_{\omega}$ which would result in a dataset with lower diversity.
On the other hand, if $\tau_{reuse}$ is set too small, then $G_{\omega}$ would be ignored in the most of the cases and the student would be following its own exploration policy.
After all these steps, if $a_t$ is still non-determined, the student follows its own policy and decide $a_t$ by $\pi_{S}(a \mid s_{t})$.

\begin{algorithm}[t]
\caption{Action Advising with Advice Imitation}
\label{alg:proposed_approach}
\begin{algorithmic}[1]
    \STATE {\bfseries Input:}     
    action advising budget $b$, 
    student policy $\pi_S$, 
    teacher policy $\pi_{T}$, 
    number of training iterations $t_{max}$,
    advice reuse uncertainty threshold $\tau_{reuse}$,
    advice reuse probability (episodic) $\varepsilon_{reuse}$,
    behavioural cloning variables:
    \begin{itemize}
    \item generative network $G_{\omega}$ ($G_{\omega}^{\mu}$ denotes uncertainty)
    \item dataset $D$ ($size(D)$ denotes the number of samples in $D$)
    \item dataset size to trigger training $n_{D}$
    \item number of training iterations $k_{bc}$
    \end{itemize}

    \STATE $D \leftarrow \emptyset$ \algorithmiccomment{initialise empty dataset}
    \STATE \emph{reuse allowed} $\leftarrow False$ \algorithmiccomment{set advice reuse off by default}
    
    \FOR{training steps $t \in \{1, 2, \ldots t_{max}\}$}
    
        \IF{$size(D) == n_{D}$}
            \STATE Train $G_{\omega}$ for $k_{bc}$ iterations \algorithmiccomment{behavioural cloning}
        \ENDIF

        \STATE $a_t \leftarrow None$ \algorithmiccomment{set action as non-determined}
        
        \IF{$Env$ is reset}
            \STATE \(u \sim \mathcal{U}(0, 1)\) \algorithmiccomment{draw a number uniformly at random}
            \IF{\(u < \varepsilon_{reuse}\)}
                \STATE \emph{reuse allowed} $\leftarrow True$
            \ELSE
                \STATE \emph{reuse allowed} $\leftarrow False$
            \ENDIF
        \ENDIF
        
        \STATE get observation $s_t \sim Env$ if $Env$ is reset
     
        \IF{$b > 0$}
            \STATE $a_{t} \sim \pi_{T}$ \algorithmiccomment{collect advice}
            \STATE add $\langle s_t, a_{t}\rangle$ to $D$
            \STATE $b \leftarrow b - 1$ \algorithmiccomment{decrement budget by 1}
        \ENDIF
        
        \IF{$a_{t}$ is $None$ {\bf and} $\pi_{S}(a \mid s_{t})$ is explorative {\bf and}\\ 
        \quad $G_{\omega}$ is trained {\bf and} $G_{\omega}^{\mu}(s_t) < \tau_{reuse}$ {\bf and}\\
        \quad \emph{reuse allowed}} 
            \STATE $a_{t} \leftarrow \argmax{a} G_{\omega}(a \mid s_t)$ \algorithmiccomment{generate imitated advice}
        \ENDIF
        
        \IF{$a_t$ is $None$}
            \STATE $a_t \sim \pi_{S}$ \algorithmiccomment{e.g., epsilon-greedy}
        \ENDIF 
        \STATE Execute $a_t$ and obtain $r_t$, $s_{t+1} \sim Env$    
        \STATE Update task-level model, e.g., DQN.
        \STATE $s_t \leftarrow s_{t+1}$
    \ENDFOR
\end{algorithmic}
\end{algorithm}

\section{Evaluation Domain} \label{sec:evaluation_domain}

\begin{figure}[t]
\centering
\begin{subfigure}[b]{0.26\linewidth} 
    \centering
    \setlength{\fboxsep}{0pt}\fbox{\includegraphics[width=0.85\textwidth]{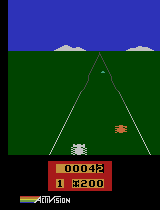}}
    \caption{Enduro}
\end{subfigure}
\begin{subfigure}[b]{0.26\linewidth} 
    \centering
    \setlength{\fboxsep}{0pt}\fbox{\includegraphics[width=0.85\textwidth]{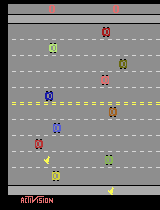}}
    \caption{Freeway}
\end{subfigure}
\begin{subfigure}[b]{0.26\linewidth}
    \centering
    \setlength{\fboxsep}{0pt}\fbox{\includegraphics[width=0.85\textwidth]{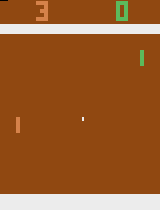}}
    \caption{Pong}
\end{subfigure}
\caption{Screenshots from the games of Enduro (a), Freeway (b) and Pong (c) within the Arcade Learning Environment.}
\label{fig_envs}
\end{figure}

In order to have a significant complexity level as well as the challenges that are relevant to the deep RL methods in our experiments, we chose three Atari 2600 games from the commonly used Arcade Learning Environment (ALE) \cite{DBLP:journals/jair/BellemareNVB13} as our evaluation domain:

\begin{itemize}
    \item \textbf{Enduro:} 
    The player controls a racing car in a long-distance track over multiple in-game days. 
    In each day, if the player manages to pass a certain number of other cars ($200$ in the first day, $300$ in the rest) in the race, it gets to advance to the next day.
    Progression during the days is visualised by different colour schemes that resemble the day-night cycle.
    Furthermore, there are other factors of seasonal events that affect the gameplay such as fogs and icy patches appearing on the road.
    Finally, as the days progress, the game increases in difficulty due to the other cars' behaviour becoming more aggressive.

    \item \textbf{Freeway:}
    In this game, the objective is to cross a chicken across a highway comprised of ten lanes with vehicles traversing in different directions and speeds.
    If the player hits the cars along the way, it gets pushed back towards starting point.
    Every time the player manages to reach the goal, it acquires a reward and gets teleported back at the starting point.
 
    \item \textbf{Pong:}
    This game consists of two paddles on the each side of the screen and a ball traversing around.
    The paddles are controlled by one player each.
    The players must hit the incoming balls to avoid them passing through their side as well as getting them thrown back at the opponent.
    If a player lets the ball pass through the gap behind its paddle, the opponent earns $1$ point.
    In the single-agent variant of this game used in our study, the player controls the right side paddle while the other one is controlled by a built-in AI.
\end{itemize}

Each of these games has an observation size of $160 \times 210 \times 3$, representing RGB images of the game screen that are produced at $60$ frames per second (FPS).
To make experimenting in these games computationally tractable, we employ some preprocessing steps that are also followed commonly in other studies \cite{DBLP:journals/corr/abs-1812-06110}.
First of all, each observation is made greyscale and resized down to the size of $80 \times 80 \times 1$.
Since the games run at a high FPS, the frame that is shown to the player is set to be only every $4$th one (which is composed of the maximum pixel values of previous $3$ frames), and the player's actions are repeated for the skipped frames.
Moreover, since these games contain a fair amount of partial observability, such as the direction of the ball in Pong, the final form of the observation to be perceived by the player is made to be a stack of $4$ pre-processed frames with a size of $84 \times 84 \times 4$ (which contains the information of the most recent $16$ actual game frames).
In order to deal with the varying range of reward scales and reward mechanisms within these games, every reward obtained in a single step in the game is clipped to be in $[-1, 1]$.
Finally, every game episode is limited to last for maximum $108$k frames, which corresponds to approximately $30$ minutes of actual gameplay time in real-life.

Another set of modifications also take place to introduce more stochasticity within the games to turn them into more challenging RL tasks.
In the beginning of the games, the player takes no-op actions for a random number of times in $[0, 30]$, to simulate the effect of having different initial states.
Additionally, with a probability of $0.25$, the actions executed by the player are repeated for an additional step, which is referred to as \emph{sticky actions}.
\section{Experimental Setup} \label{sec:experimental_setup}

The goal of our experiments\footnote{\text{Code for our experiments can be found at }\url{https://github.com/ercumentilhan/naive-advice-imitation}} is to demonstrate that it is possible to generalise the teacher advice to the unseen yet similar states with our method, and that it is an effective way of improving performance of action advising, in complex domains especially.
Therefore we choose the games described in Section~\ref{sec:evaluation_domain} as our test-beds.
The set of the student agent variants we compare are listed as follows:

\begin{itemize}
	\item \textbf{No Advising (None)}: 
	No action advising procedure is followed; the student learns as normal.
	
	\item \textbf{Early Advising (EA)}: 
	The student follows early advising heuristic to distribute its advising budget.
	Specifically, the teacher is queried for an advice at every step until the budget runs out.
	
	\item \textbf{Early Advising with Advice Reuse via Imitation (AR)}: 
	The student follows our proposed strategy (Section~\ref{sec:proposed_approach}) combined with early advising heuristic.
	It starts off by greedily asking for advice until its budget runs out; then, it activates its behavioural cloning module to imitate and reuse the previously collected advice in the remaining exploration steps.

\end{itemize}

All student agent variants employ the identical task-level RL algorithm which is DQN with double Q-learning and dueling networks enhancements, and $\epsilon$-greedy policy as the exploration strategy.
The convolutional neural network structure within the DQN in input-to-output order is as follows: $32$ $8 \times 8$ filters with a stride of $4$, $64$ $4 \times 4$ filters with a stride of $2$, $64$ $3 \times 3$ filters with a stride of $1$, followed by a fully-connected layer with $512$ hidden units and multiple streams that add up in the end (dueling).
Additionally, the student agent variant AR also incorporates a behaviour cloning module, which is a neural network with an identical structure minus the dueling stream.
All the layer activations are set to be ReLU.
The hyperparameters are tuned prior to experiments and kept the same across all experiments can be seen in Table~\ref{table:hyperparameters}.

In this teacher-student setup, we also need a teacher from which the student can get good quality action advice.
For this purpose, we trained a DQN agent for each of these games for $10$M steps ($40$M actual game frames) to achieve a competent level performance in each.

The experiments are conducted by executing every student variant through a learning session $3$M steps ($12$M actual game frames) for every game.
The learning steps are kept relatively small compared to the teacher training since it is expected for the students to achieve high performance much quicker with the aid of advice.
Through the learning sessions, the agents are also evaluated at every $25$k$^{th}$ step in a separate instance of the environment for $10$ episodes.
During evaluation, any form of exploration and teaching is disabled in order to assess the actual proficiency of the students.

In terms of action advising setup, we set the action advising budget as $10$k steps which corresponds to only approximately $0.3\%$ of the interactions in a learning session and also to almost one third of a full game episode ($27$k steps).
Besides the budget, our proposed method AR also uses some additional hyperparameters which were tuned prior to the full length experiments and are kept the same across every game.
The dataset size $n_{D}$ to train $G_{\omega}$ is set as $10$k which is the action advising budget as we employ early advising prior to behavioural cloning training.
The number iterations to train $G_{\omega}$ is set as $50$k.
Episodic advice reuse probability $\varepsilon_{reuse}$ is set as $0.5$ meaning that the student will follow $G_{\omega}$ in half the episodes (in the appropriate states).
Finally, advice reuse uncertainty threshold $\tau_{reuse}$ is set as $0.01$ (determined empirically) and kept the same across all games.
In the experiments with AR, we also record the actual advice actions generated by the teacher at every step (not seen by the student) to have access to the ground-truth values to measure the accuracy of the behavioural cloning module.
Every particular experiment case is repeated and aggregated over $3$ different random seeds.

\begin{table}[!t]
	\centering
	\caption{Hyperparameters used in the student's DQN (top section) and Behaviour Cloning Network (bottom section).}
	\label{table:hyperparameters}
	\begin{tabular}{l|c}  	
		Hyperparameter name & Value \\
  	    \cmidrule(r){1-2}
  	    Replay memory initial size and capacity & $50$k, $500$k \\
		Target network update period & $7500$ \\
		Minibatch size & $32$ \\
		Learning rate & $625 \times 10^{-7}$ \\
		Train period & $4$ \\
		Discount factor $\gamma$ & $0.99$ \\
		$\epsilon$ initial, $\epsilon$ final, $\epsilon$ decay steps & $1.0$, $0.01$, $500$k \\
		\cmidrule(r){1-2}
		Minibatch size & $32$ \\
		Learning rate & $0.0001$ \\
		Dropout rate & $0.2$ \\
		\# of forward passes to assess uncertainty & $100$ \\

	\end{tabular}
\end{table}

\section{Results and Discussion} \label{sec:results_and_discussion}

\begin{figure*}[!t]
\centering
\includegraphics[width=0.9\textwidth, keepaspectratio]{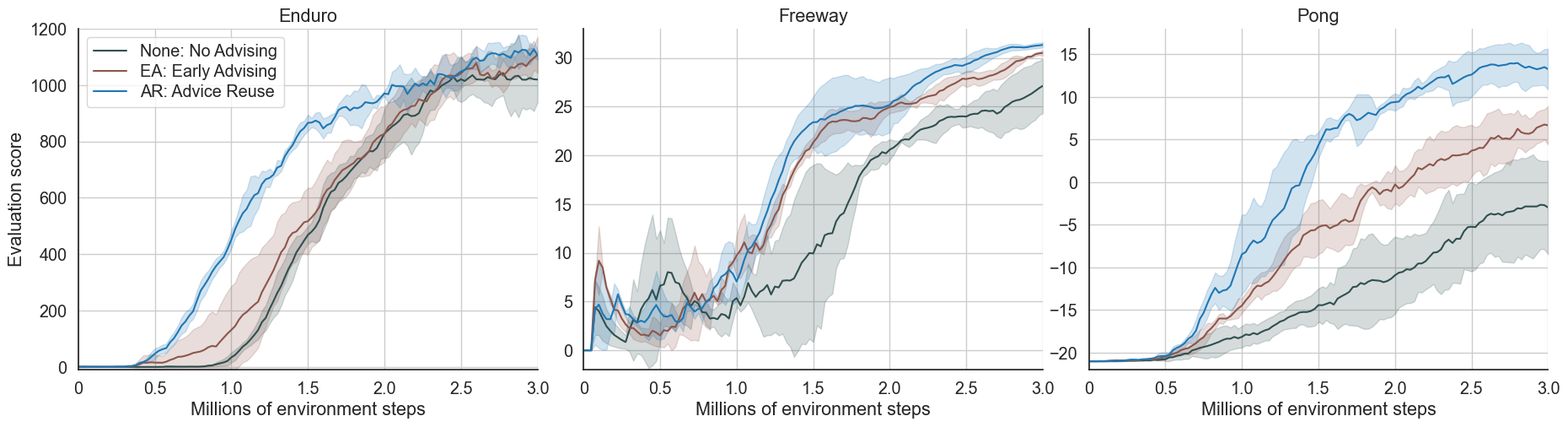}
\caption{
Evaluation scores of the student variants None, EA, AR obtained in the Atari games of Enduro (leftmost column), Freeway (middle column), Pong (rightmost column) aggregated over $3$ runs.
Shaded areas show the standard deviation across the runs.
}
\label{fig:results_eval}
\end{figure*}

\begin{figure*}[!t]
\centering
\includegraphics[width=0.9\textwidth, keepaspectratio]{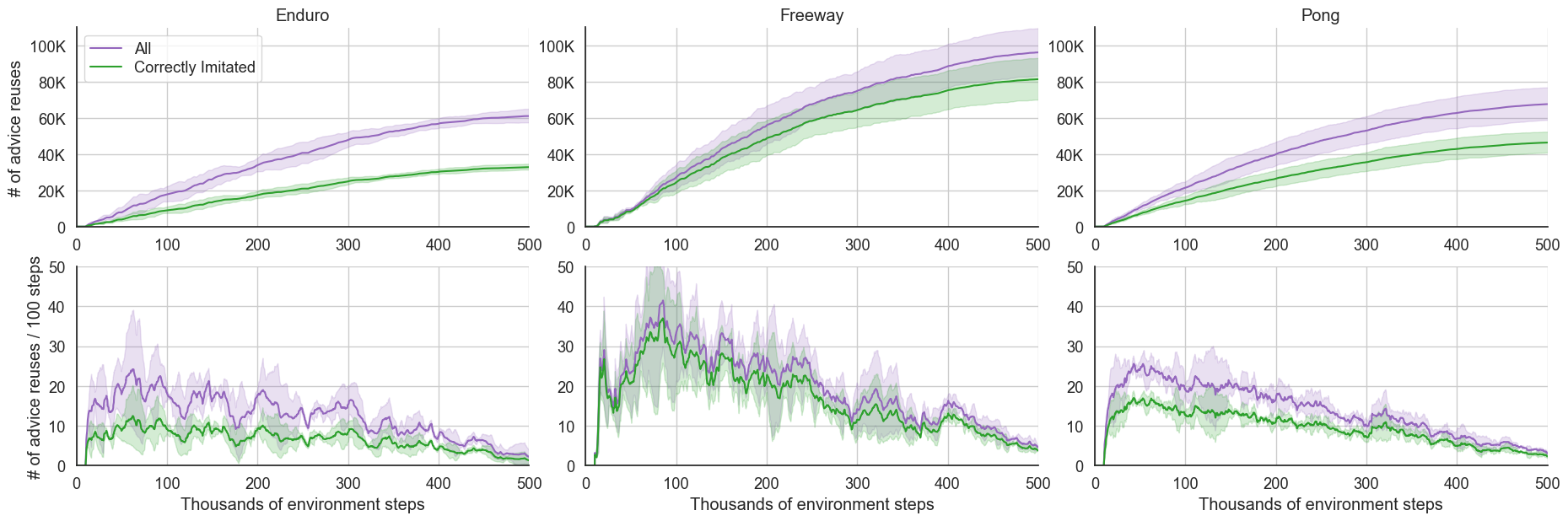}
\caption{
Number of advice reuses performed by the student with AR mode in the Atari games of Enduro (leftmost column), Freeway (middle column), Pong (rightmost column) over $3$ runs, plotted cumulatively (top row) and in every $100$ steps (bottom row).
Purple lines represent the number of all advice reuses while the green lines represent the number of correctly imitated ones among these.
Shaded areas show the standard deviation across the runs.
}
\label{fig:results_advices}
\end{figure*}

\begin{table*}[t]
	\centering
	\caption{
	Final and area-under-the-curve (AUC) values of evaluation score plots (Figure~\ref{fig:results_eval}), the number of exploration steps, the number of advice reuses (all and correctly imitated) of None, EA, AR student modes obtained in the Atari games of Enduro, Freeway, Pong aggregated over $3$ runs.
	The numbers denoted by $\pm$ indicate standard deviation.
	The numbers in the parentheses show the percentage of reused advices in the exploration steps (in the column titled ``All'' ) and the percentage of correctly imitated advices in total number of reused advices (in the column titled ``Correctly Imitated'' ).
	}
	\label{table:results}
	\begin{tabular}{clccccc}  
	\toprule
	& & \multicolumn{2}{c}{Evaluation Score} & \multirow{2}{*}{\# of Exp. Steps} & \multicolumn{2}{c}{\# of Advice Reuses} \\
	\cmidrule(r){3-4}
	\cmidrule(r){6-7}
	Game & Mode & \multicolumn{1}{c}{Final} & \multicolumn{1}{c}{AUC ($\times 10^2$)} & 
	& \multicolumn{1}{c}{All} & \multicolumn{1}{c}{Correctly Imitated} \\	
	\midrule
	
	\multirow{3}{*}{Enduro} &
	None &
	$1021.54 \pm 79.5$ & $570.61 \pm 38.4$ 
	& 
	$326939 \pm 92.1$
	& 
	{---} & {---} 
	\\
   	& EA &
	$1095.55 \pm 45.9$ & $616.29 \pm 58.1$
	& 
	$326753 \pm 220.9$
	& 
	{---} & {---} 
    \\
	& AR &
	$\bm{1112.79 \pm 16.6}$ & $\bm{782.98 \pm 8.4}$
	& 
	$326889 \pm 230.5$
	& 
	$67198 \pm 3061.0$ $(20.55\%)$ & $36534 \pm 1210.9$ $(54.44\%)$ 
    \\
    
    \cmidrule(r){1-7}

	\multirow{3}{*}{Freeway} &
	None &
	$26.87 \pm 2.3$ & $15.73 \pm 1.7$
	& 
	$326872 \pm 199.9$
	& 
	{---} & {---}
	\\
   	& EA &
	$30.44 \pm 0.2$ & $20.31 \pm 0.4$ 
	& 
	$327158 \pm 6.2$
	& 
	{---} & {---}
    \\
	& AR &
	$\bm{31.28 \pm 0.2}$ & $\bm{21.52 \pm 1.0}$ 
	& 
	$326778 \pm 494.4$
	& 
	$104770 \pm 12522.2$ $(32.05\%)$ & $88829 \pm 10950.5$ $(84.74\%)$
    \\

    \cmidrule(r){1-7} 
	\multirow{3}{*}{Pong} &
	None &
	$-2.78 \pm 4.3$ & $-16.24 \pm 2.6$
	& 
	$326744 \pm 25.2$
	& 
	{---} & {---} 
	\\
   	& EA &
	$6.66 \pm 1.6$ & $-8.83 \pm 0.4$ 
	& 
	$326872 \pm 199.9$
	& 
	{---} & {---} 
    \\
	& AR &
	$\bm{13.35 \pm 1.7}$ & $\bm{-1.36 \pm 1.0}$
	& 
	$326933 \pm 371.2$
	& 
	$72581 \pm 7615.7$ $(22.20\%)$ & $49538 \pm 4853.8$ $(68.32\%)$ 
    \\

	\bottomrule
	\end{tabular}
\end{table*}

\label{fig:results_eval}
\label{fig:results_advices}
\label{table:results}

The results of our experiments are presented in Figures~\ref{fig:results_eval},\ref{fig:results_advices} and Table~\ref{table:results}.
Figure~\ref{fig:results_eval} contains the plots for the evaluation scores obtained by None, EA and AR modes of the student in the games of Enduro, Freeway, Pong.
In Figure~\ref{fig:results_advices}, the plots of the advice reuse trends of AR in this set of games are displayed as cumulatively (top row) and in every $100$ steps windows (bottom row).
These plots are limited to the first $500$k steps to only consider the exploration stage determined by the agent's $\epsilon$-greedy schedule.
Purple lines here represent all advice reuses combined, while the green lines indicate only the correctly imitated (in terms of being equal to the ground-truth teacher advice) advice pieces.
These results are also reported in Table~\ref{table:results} in the numerical form where the evaluation scores are broken down in two parts of final value and area-under-the-curve, which represent the final agent performance and the learning speed, respectively.
Furthermore, the table also contains the total number of exploration steps taken, as well as the percentage of the number of reused advice in the exploration steps and the percentage of correctly imitated advice in total number of reused advice (denoted in parentheses).

In the evaluation scores, we see different outcomes in each of these games.
In Enduro, we see that AR provides a significant amount of jump start and performs the best in terms of learning speed while being far ahead of EA and None which are quite similar.
When it comes to the final performance however, while EA and AR both outperform None, they do not differ much from each other.
In Freeway, EA and AR perform very similarly in terms of learning speed and final performance with AR being slightly ahead of EA.
However, they outperform None significantly.
This shows that it matters to be advised initially, though their repetitions may not always yield much acceleration in learning.
Finally, in Pong, we see a great difference between the performances in every aspect.
Our AR comes out far ahead than its closest follower EA both in terms of final score and learning speed.
This is an example of how getting a very little advice in the beginning as well as repeating them across further explorative actions can cause a great impact on the learning.
Overall, AR manages to be the best in every game and suffers no performance loss even with high advice utilisation (as high as ~$104$k in Freeway) which was shown to be harmful to learning in previous studies.
Even though its performance boost over None seems to be not huge in every scenario, it should be noted that this is the case of it being combined with EA baseline.
With more complicated methods, AR can be capable of training its imitation learning module with a more diverse set of experience and therefore, have a larger coverage which can potentially yield superior performance.

The task-level performance of our approach is affected primarily by two factors: the accuracy of advice imitation and its coverage/usage in the remainder of the exploration steps (the process of reusing).
Therefore, we also analyse the advice reuse statistics of AR to form links between these outcomes.
First of all, it should be noted that the decreasing trend in these plots is caused by the $\epsilon$-greedy annealing.
Enduro is the game with the smallest advice reuse rate as well as the lowest imitation accuracy.
This is possibly because of the game episodes lasting long regardless of the agent's performance, which is likely to reduce the proportion of the familiar states according to the behavioural cloner.
In Freeway, we observe a fairly high advice reuse rate with high accuracy of imitation.
However, this is not reflected in the performance difference obtained versus EA, unlike in Enduro and Pong.
Finally, in Pong, where the performance improvement is the most significant, advice reuse ratio seems to be similar to Enduro, but with far higher imitation accuracy.

Clearly, as we see from all these results combined, we can say that it is definitely a viable idea to extend the teacher advice over future states through imitation since this can be achieved with relatively high accuracy.
However, even when we have access to these imitated competent policies, it is still non-trivial to construct a \emph{good} exploration policy.
While a higher advice reuse rate produces a more consistent exploration policy with less random dithering, it also has the risk of limiting the sample diversity in the replay memory, which can be problematic especially if the imitation quality is also poor.
As long as the reuse amount does not get excessively high, it is safe to have the imitation learning accuracy around these reported levels, which makes tuning the uncertainty threshold straightforward.
This is especially important for the realistic applications where it is not possible to access the tasks to tune such hyperparameters beforehand.

\balance
Finally, we also analyse our approach's computational burden, which may be the primary concern when adopting it.
Specifically, it involves two extra operations: behavioural cloning network training and uncertainty estimations.
The former happens only once in the beginning and therefore is negligible.
The uncertainty estimations that require multiple forward passes (which is $100$ in our experiments) happens in every exploration step and was found to cause a maximum of $2\times$ slowdown in our experiments.
Considering that the exploration steps only spans approximately $10\%$ of a learning session, we can expect the runs to be taking at most $10\%$ longer in total when AR is employed in a similar setting to ours; and, this becomes even smaller when the learning sessions last longer in terms of the total number of environment steps.
Clearly, this is a small setback considering the sample efficiency benefits our method brings.

\section{Conclusions and Future Work} \label{sec:conclusions_and_future_work}

In this study, we developed an approach for the student to imitate and reuse advice previously collected from the teacher.  
This is the first time such an approach has been proposed in deep reinforcement learning (RL).
In order to do so, we followed an idea similar to behavioural cloning, employing a separate neural network that is trained with the advised state-action pairs via supervised learning.
Thus, this module can imitate the teacher's policy in a generalisable way that lets us apply it to the unseen states.
We also incorporated a notion of epistemic uncertainty via dropout in this neural network to be able to limit the imitations to the states that are similar to the advice collected states.

The results of the experiments in $3$ Atari games have shown that it is a feasible idea to  accurately generalise a small set of teacher advice over unseen yet similar states in future.
Furthermore, our approach of employing behavioural cloning was found to be a successful way of achieving this, as it yielded a considerably high accuracy of imitation in multiple games.
Additionally, reusing these self-generated advice across the exploration steps provided significant improvements in the learning speeds and the final performances without any over-advising induced performance deterioration.
Therefore, our method can be considered as a promising enhancement to the existing action advising methods, especially since it is also very straightforward to implement and tune, with only a small computational burden.
Finally, it was also seen that utilisation of such imitated advice policies to construct good quality exploration is non-trivial and requires further investigation.

Our study lies at the intersection of action advising and exploration in RL and can be extended in various interesting ways.
It is unclear how far the different qualities of imitation and reuse rates can affect performance in one particular game; it will be a worthwhile study to analyse these.
Furthermore, evaluating the advice in terms of its contribution to learning progress is a promising direction to take.

\clearpage
\bibliographystyle{ACM-Reference-Format} 
\bibliography{bibdb}

\end{document}